\documentclass[letterpaper,journal]{IEEEtran}
\usepackage{amsmath,amsfonts}
\usepackage{array}
\usepackage{textcomp}
\usepackage{stfloats}
\usepackage{url}
\usepackage{verbatim}
\usepackage{graphicx}
\usepackage{cite}
\usepackage{xcolor}
\usepackage{colortbl}
\usepackage{subcaption}
\usepackage{mathtools}  
\usepackage{amssymb}
\usepackage{amsthm}
\usepackage{tabulary}
\usepackage{booktabs}
\usepackage[ruled,vlined]{algorithm2e}
\usepackage{algorithmic}
\usepackage{hyperref}
\usepackage{setspace}
\usepackage{subfiles}
\usepackage{multirow}
\usepackage{bm}
\usepackage{makecell}
\allowdisplaybreaks

\newcommand{\RN}[1]{\uppercase\expandafter{\romannumeral#1\relax}}

\DeclareMathOperator*{\argmax}{arg\,max}

\newcommand{\df}[1]{\mathrm{d}{#1}}

\title{

  ATRS: Adaptive Trajectory Re-splitting via a Shared Neural Policy for Parallel Optimization

}

\author{
        \IEEEauthorblockN{
          {\small
        Jiajun Yu,
        Guodong Liu, 
        Li Wang,
        Pengxiang Zhou,
        Wentao Liu,
        Yin He,
        Chao Xu,
        Fei Gao,
        and Yanjun Cao\textsuperscript{*}
          }
        }
        \vspace{-1.0cm}
        }

\begin{document}

\maketitle
\begin{abstract}
  Parallel trajectory optimization via the Alternating Direction Method of Multipliers (ADMM) has emerged as a scalable approach to long-horizon motion planning.
  However, existing frameworks typically decompose the problem into parallel subproblems based on a predefined fixed structure.
  Such structural rigidity often causes optimization stagnation in highly constrained regions, where a few lagging subproblems delay global convergence.
  A natural remedy is to adaptively re-split these stagnating segments online.
  Yet, deciding when, where, and how to split exceeds the capability of rule-based heuristics.
  To this end, we propose \textbf{ATRS}, a novel framework that embeds a shared Deep Reinforcement Learning policy into the parallel ADMM loop.
  We formulate this adaptive adjustment as a Multi-Agent Shared-Policy Markov Decision Process, where all trajectory segments act as homogeneous agents and share a unified neural policy network.
  This parameter-sharing architecture endows the system with size invariance, enabling it to handle dynamically changing segment counts during re-splitting and generalize to arbitrary trajectory lengths.
  Furthermore, our formulation inherently supports zero-shot generalization to unseen environments, as our network relies solely on the internal states of the numerical solver rather than on the geometric features of the environment.
  To ensure solver stability, a Confidence-Based Election mechanism selects only the most stagnating segment for re-splitting at each step.
  Extensive simulations demonstrate that ATRS accelerates convergence, reducing the number of iterations by up to 26.0\% and the computation time by up to 19.1\%.
  Real-world experiments further confirm its applicability to both large-scale offline global planning and real-time onboard replanning within 35\,ms per cycle, with no sim-to-real degradation.
\end{abstract}


\section{Introduction}
\IEEEPARstart{T}{rajectory} optimization is essential for autonomous robots to navigate safely and efficiently in complex environments.
As applications expand to large-scale environments and swarm-level tasks \cite{environmental_monitoring, dong2025multi}, optimizing long-horizon trajectories in real time becomes a critical computational challenge.
Trajectory optimization is commonly formulated as an Optimal Control Problem (OCP).
General-purpose solvers like GPOPS-II \cite{GPOPS-II} can handle rich constraint formulations with high numerical accuracy. 
However, their dense Hessian matrix leads to cubic time complexity $O(N^3)$, rendering them unsuitable for real-time long-horizon planning.
To mitigate this, recent solvers exploit specific problem structures to achieve linear time complexity $O(N)$. 
For example, TrajOpt \cite{trajOpt} exploits the sparsity of the Hessian matrix, while Gcopter \cite{Gcopter} leverages banded system structures.
Nevertheless, the problem formulations of these methods adopt sequential and centralized paradigms, 
thereby hindering the utilization of modern multi-core architectures for parallel acceleration.
\begin{figure}[!t]
  \centering
  \includegraphics[width=0.476\textwidth]{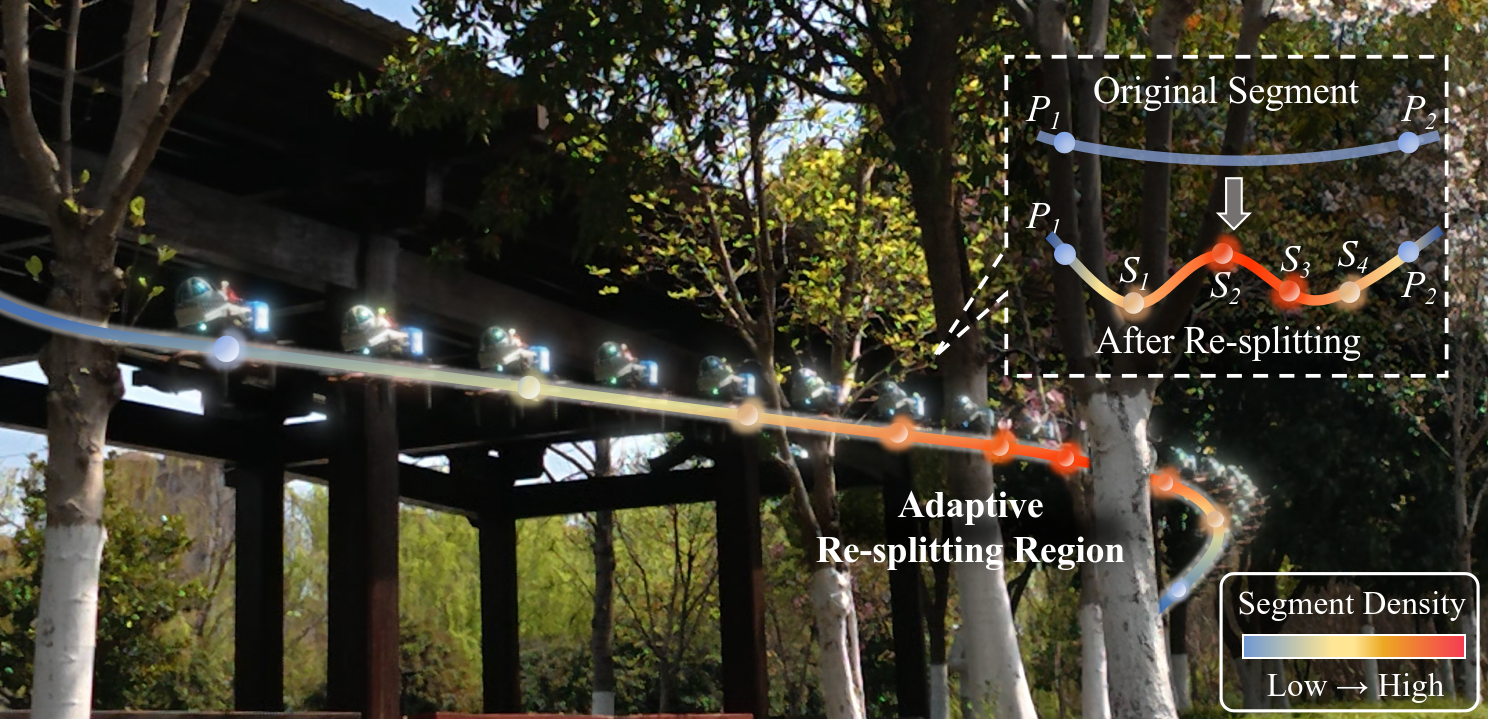}
  \caption{
  Real-world quadrotor navigation through a forest environment.
  The trajectory color encodes segment density from low (blue) to high (red).
  The inset illustrates adaptive re-splitting: the stagnating segment $P_1 P_2$ is subdivided by inserting intermediate waypoints $S_1$–$S_4$, injecting local degrees of freedom to accelerate convergence.
  }  
  \label{fig:top_figure}
  \vspace{-0.5cm}
\end{figure}

The Alternating Direction Method of Multipliers (ADMM) provides a powerful paradigm to break these computational bottlenecks through parallel optimization.
Wang \textit{et al.} \cite{trajectorySplitting} pioneered the use of Consensus ADMM (CADMM) to decompose long trajectories into multiple independent segments for parallel solving.  
Building on this, Yu \textit{et al.} proposed TOP \cite{TOP}, which integrates high-order dynamics and safety constraints, theoretically achieving constant-time complexity regardless of trajectory length.
Despite these advancements, the varying computational difficulty across these subproblems hinders parallelization from achieving ideal performance gains.
This stems from a fundamental limitation of these methods: a \textit{predefined fixed structure}.
Specifically, the number of split trajectory segments is typically pre-determined by the front-end path planner (e.g., A*\cite{D*A*}). 
However, the computational effort required for convergence varies across different segments.
Segments in open regions converge rapidly, whereas those in highly constrained areas, such as narrow safety corridors with tight dynamic constraints, require more iterations.
Since the CADMM framework relies on inter-segment consensus, these lagging subproblems inherently delay the global convergence, potentially leading to failure.


This motivates adaptive re-splitting of the trajectory segments to balance the computational load.
The system needs to jointly determine \textit{when}, \textit{where}, and \textit{how} to re-split.
Such decisions exceed the capability of handcrafted heuristics, whose fixed rules lack the adaptivity to model such complex mappings.
Recent ``Learning to Optimize'' (L2O) methods~\cite{sambharya2024learning,rlqp} offer a promising data-driven paradigm to overcome this limitation.
Existing L2O methods have achieved notable progress in parameter tuning and solver warm-starting for problems with fixed dimensions. 
However, how to adaptively reconfigure the optimization structure online despite changing problem dimensions remains an open question.

To address these challenges, we propose \textbf{ATRS} (\textbf{A}daptive \textbf{T}rajectory \textbf{R}e-splitting via a \textbf{S}hared Neural Policy), which embeds a shared Deep Reinforcement Learning (DRL) policy into the parallel ADMM loop to adaptively re-split stagnating trajectory segments.
We formulate this adaptive structural adjustment as a Multi-Agent Shared-Policy Markov Decision Process (MASP-MDP), where all trajectory segments act as homogeneous agents sharing a unified policy network.
Built on a parameter-sharing architecture, this network is \textit{size-invariant}, enabling it to handle the dynamically changing number of segments during re-splitting and to generalize to arbitrary trajectory lengths.
Crucially, the state representation relies solely on the numerical solver's internal states rather than environmental geometry.
This allows the trained policy to transfer zero-shot across unseen environments without retraining.
Furthermore, to maintain solver stability during structural changes, we devise a Confidence-Based Election mechanism, ensuring only the most stagnating segment executes a split at each step.
Extensive simulations demonstrate that ATRS reduces the iteration count by up to 26.0\% and the computation time by up to 19.1\%.
Real-world experiments further confirm its applicability to both large-scale offline global planning and real-time onboard replanning within 35\,ms per cycle, with no sim-to-real degradation (Fig.~\ref{fig:top_figure}).
The main contributions of this work are:
\begin{itemize}
    \item \textbf{Adaptive Trajectory Re-splitting Framework:}  We formulate the adaptive structural adjustment for parallel trajectory optimization as a Multi-Agent RL problem for the first time. 
    The system autonomously identifies and resolves optimization bottlenecks, preventing lagging subproblems from hindering the global convergence.
    \item \textbf{Size-Invariant Policy via Parameter Sharing:} We propose a Multi-Agent Shared Policy Network, where all trajectory segments share a unified policy.
    This renders the policy \textit{size-invariant}, enabling it to handle the dynamic segment counts during re-splitting and generalize to arbitrary trajectory lengths without retraining.

    \item \textbf{Zero-Shot Generalization and Real-World Validation:} 
    Our formulation inherently supports zero-shot generalization to unseen environments by relying solely on the internal states of the numerical solver.
	Benchmarks and real-world experiments validate that the lightweight policy enables faster convergence and real-time onboard deployment without sim-to-real degradation.

\end{itemize}

\section{Related Work}


\subsection{Parallel Trajectory Optimization}
Gradient-based trajectory optimization has become the dominant paradigm for robot motion planning due to its capacity to incorporate complex constraints \cite{kumarMinjerk, Gcopter}.
To address scalability in long-horizon trajectory generation, parallel computing architectures have been increasingly explored.
Wang \textit{et al.} \cite{trajectorySplitting} pioneered the application of CADMM to decompose the trajectory optimization problem into independent subproblems for parallel solving.
Yu \textit{et al.} \cite{TOP} extended this parallel formulation to quadrotor trajectory planning and proved that, with sufficient parallel threads, computation time can theoretically remain constant regardless of trajectory length.

However, these parallel methods rely on a fixed discretization structure determined by the front-end path planner, ignoring the unbalanced computational difficulties across segments.
Since CADMM relies on inter-segment consensus, the stagnating subproblems under tight constraints inherently delay the global convergence.
To address this, ATRS adaptively re-splits stagnating segments online, injecting local degrees of freedom to rebalance the computational load.

\vspace{-0.2cm}
\subsection{Learning-Augmented Optimization}
Existing methods for integrating neural networks with numerical optimization range from fully bypassing the solver to augmenting specific stages of its pipeline.
End-to-end methods \cite{navrl} map sensor inputs directly to control commands, bypassing the modular navigation pipeline.
However, they cannot enforce hard constraints on safety and dynamic feasibility. 
To mitigate this, hybrid architectures leverage neural networks to initialize spatio-temporal parameters \cite{neo-planner} or predict initial time allocations \cite{wu2024deep}, while relying on back-end optimization to ensure constraint satisfaction.
Whereas such hybrid methods treat the numerical solver as a fixed post-processing step, the ``Learning to Optimize'' (L2O) paradigm intervenes from within.
Recent works predict internal primal-dual variables for warm-starting \cite{sambharya2024learning} or dynamically tune penalty parameters ($\rho$) \cite{rlqp} to improve convergence rates.
Despite these solver-level advances, existing L2O methods remain confined to parameter tuning or warm-starting, leaving the core problem structure unchanged. 
For segments stagnating under tight constraints, adjusting solver parameters alone cannot resolve the underlying structural rigidity.
Furthermore, existing neural architectures (e.g., \cite{wu2024deep}) rely on fixed maximum input dimensions, severely limiting generalization to arbitrary trajectory lengths.
Alternatively, sequence models like Transformers can accommodate variable dimensions in theory, but their computational overhead makes them impractical for real-time onboard deployment.
In contrast, ATRS advances the L2O paradigm from learning solver parameters to learning problem structure.
It provides a lightweight, size-invariant framework that generalizes to arbitrary trajectory lengths.
\section{Methodology}

In this section, we present the proposed adaptive trajectory re-splitting framework, {ATRS}. 
The underlying trajectory optimization is first formulated as a consensus parallel optimization problem via CADMM (Sec. \ref{Sec3A:preliminaries}). 
To address the structural rigidity of the solver, Sec. \ref{Sec3B:mdp_formulation} models the adaptive trajectory reconfiguration as a MASP-MDP, including the state representation, action space, and reward design.
Sec. \ref{Sec3C:policy design} then introduces the algorithm selection and the architecture of the shared policy network.
Next, we introduce a Confidence-Based Election mechanism to ensure solver stability (Sec. \ref{Sec3D:vote}).
Finally, Sec. \ref{Sec3E:train} summarizes the training and deployment pipeline of ATRS.

\begin{figure}
  \centering
  \includegraphics[width=0.476\textwidth]{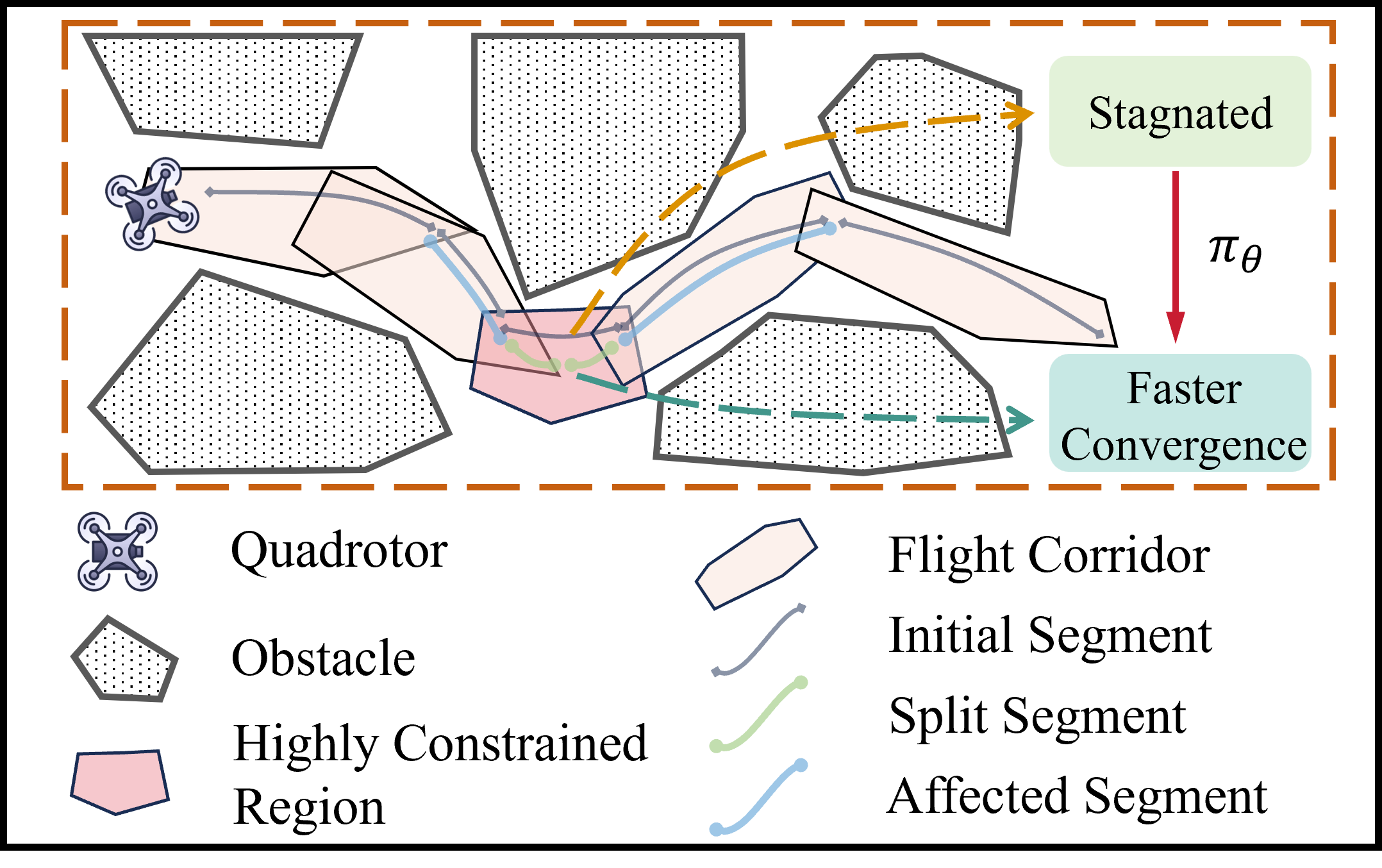}
  \caption{Illustration of the re-splitting mechanism.
  When a trajectory segment stagnates in a highly constrained region (pink), the policy $\pi_\theta$ triggers a re-splitting. 
  The newly introduced sub-segments (green) inject local degrees of freedom, enabling faster convergence.}
  \label{fig:framework}
  \vspace{-0.5cm}
\end{figure}

\vspace{-0.3cm}
\subsection{Preliminaries}
\label{Sec3A:preliminaries}
The trajectory optimization problem aims to find a smooth trajectory that minimizes control effort subject to safety and dynamic constraints.
By leveraging the differential flatness property of quadrotors, we can analytically recover the full system state and control inputs from a lower-dimensional flat output $\boldsymbol{\sigma}(t) \in \mathbb{R}^m$.
Following the formulation in TOP \cite{TOP}, we parameterize the global trajectory as a piecewise polynomial consisting of $N$ segments, where $N$ is initially determined by the front-end path planner.
To solve this problem efficiently in parallel, we adopt the CADMM framework, where continuity between adjacent segments is enforced via shared consensus variables. 
The distributed optimization problem is formulated as follows:
\begin{equation}
  \begin{aligned}
    \label{eq:piecewise_optimization}
    \min_{\boldsymbol{\sigma}(t)} ~& \sum_{i=1}^{N} \int_{0}^{T_i} \boldsymbol{\sigma}_i^{(p)}(t)^\top \mathbf{W} \boldsymbol{\sigma}_i^{(p)}(t) \df{t}, \\
    s.t. ~& \widetilde{\boldsymbol{\sigma}}_i^{[d-1]} = \widetilde{\mathbf{z}}_i,~i \in \{1,\dots, N\}, \\
    & ~ \mathcal{G}(\boldsymbol{\sigma}(t),\dot{\boldsymbol{\sigma}}(t),\ldots,\boldsymbol{\sigma}^{(p)}(t))\preceq\mathbf{0},~\forall t \in [0,T_i],
  \end{aligned}
\end{equation}
where $\boldsymbol{\sigma}_i^{(p)}(t)$ denotes the $p$-th order derivative (control input) of the $i$-th segment, weighted by the diagonal matrix $\mathbf{W} \in \mathbb{R}^{m \times m}$.
The boundary state vector is defined as $\widetilde{\boldsymbol{\sigma}}_i^{[d-1]} \triangleq [(\boldsymbol{\sigma}_i^{[d-1]}(0))^\top, (\boldsymbol{\sigma}_i^{[d-1]}(T_i))^\top]^\top$, where $\boldsymbol{\sigma}^{[n]} \triangleq [\boldsymbol{\sigma}^\top, \dot{\boldsymbol{\sigma}}^\top, \ldots, (\boldsymbol{\sigma}^{(n)})^\top]^\top$ represents the chain of integrators.
Correspondingly, $\widetilde{\mathbf{z}}_i \triangleq [\mathbf{z}_{i-1}^\top, \mathbf{z}_i^\top]^\top$ enforces the boundary conditions, where $\mathbf{z}_i \in \mathbb{R}^{md}$ represents the consensus variable at the interface between the $i$-th and $(i+1)$-th segments (including derivatives up to order $d-1$).
The function $\mathcal{G}(\cdot)$ imposes convex constraints that guarantee dynamic feasibility and collision avoidance.

In Eq. (\ref{eq:piecewise_optimization}), the segment count $N$ remains fixed throughout the optimization.
However, in complex environments, this rigid structure leads to unbalanced convergence across segments. 
Those under tight constraints require more iterations, while others converge rapidly.
To resolve it, we propose to adaptively re-split the stagnating segments, injecting local degrees of freedom through targeted structural changes (Fig.~\ref{fig:framework}).
Owing to CADMM's parallel architecture, the newly introduced segments do not increase the per-iteration time complexity.


\subsection{MASP-MDP Formulation}
\label{Sec3B:mdp_formulation}
Unlike prior L2O methods that tune solver parameters, our formulation directly learns to adjust the problem structure.
Since the number of segments $N$ varies dynamically during the re-splitting process, standard fixed-input neural networks are inapplicable.
To address this, we formulate the structural adaptation problem as a Multi-Agent Shared-Policy Markov Decision Process (MASP-MDP).
In this framework, the trajectory optimization problem serves as the environment, and each trajectory segment acts as a homogeneous agent $i \in \{1, \dots, N\}$. 
Instead of training distinct policies for each agent, all agents share a unified policy network $\pi_\theta$.
This formulation adapts the framework of Huang \textit{et al.} \cite{huang2020one} and can be viewed as a specialized instance of a Multi-Agent MDP \cite{lowe2017multi} or a Markov game \cite{littman1994markov}.
The process is defined by the tuple $(\mathcal{S}, \mathcal{A}, \mathcal{R}, \gamma)$, where $\mathcal{S}$ is the state space, $\mathcal{A}$ is the action space, $\mathcal{R}: \mathcal{S} \times \mathcal{A} \rightarrow \mathbb{R}$ is the reward function, and $\gamma \in [0,1)$ is the discount factor.
The goal is to learn an optimal policy $\pi^*$ that maximizes the expected cumulative discounted reward over an episode of horizon $T$:
\begin{equation}
\label{eq:rl_objective}
\pi^* = \argmax_{\pi} \mathbb{E} \left[ \sum_{t=0}^{T} \gamma^t R(s_t, a_t) \right].
\end{equation}

\subsubsection{State Space}
To enable agents to perceive local optimization status while maintaining awareness of global solver trends, we design a composite observation vector $\mathbf{s}_i \in \mathbb{R}^{11}$. 
This vector comprises both local and global features:
\begin{equation}
\mathbf{s}_i = \left[ (\mathbf{s}_i^{\text{loc}})^\top, (\mathbf{s}^{\text{glo}})^\top \right]^\top,
\end{equation}
where the local state $\mathbf{s}_i^{\text{loc}} \in \mathbb{R}^6$ encodes both the solver's internal numerical metrics and the spatio-temporal properties of the $i$-th segment, while the global state $\mathbf{s}^{\text{glo}} \in \mathbb{R}^5$ provides the overall solver status, shared across all agents.


\textbf{Local State ($\mathbf{s}_i^{\text{loc}}$):}
The local state characterizes the optimization condition of the corresponding trajectory segment. 
It fuses the ADMM solver's internal convergence metrics with spatio-temporal trajectory properties.
Given the wide dynamic range of internal numerical metrics (often ranging from $10^{-6}$ to $10^{6}$), we apply a logarithmic mapping to the magnitude-based features to ensure numerical stability during training.
The local state is defined as 
\begin{equation}
\mathbf{s}_i^{\text{loc}} = \left[~ \log_{10}(\epsilon_i), ~\log_{10}(\|\mathbf{y}_i\|), ~\bar{J}_i, ~\tau_{\text{trend}}, ~T_i, ~\delta_i~ \right]^\top,
\end{equation} 
where each component is defined as follows:
\begin{itemize}
  \item Total Residual ($\epsilon_i$):
  Defined as $\epsilon_i = \|\mathbf{r}_i^p\|^2 + \|\mathbf{r}_i^d\|^2$, where $\mathbf{r}_i^p = \widetilde{\boldsymbol{\sigma}}_i^{[d-1]} - \widetilde{\mathbf{z}}_i$ is the primal residual measuring the consensus violation in Eq. (\ref{eq:piecewise_optimization}), and $\mathbf{r}_i^d$ is the corresponding dual residual.
  A high value indicates severe constraint violations, highlighting potential stagnation.

  \item Dual Variable Norm ($\|\mathbf{y}_i\|$): 
  The $L_2$-norm of the continuity dual variables. 
  A persistently high and growing norm indicates accumulated constraint stress and often precedes optimization stagnation.

  \item Energy Density ($\bar{J}_i$): 
  Computed as $\log_{10}(J_i / T_i)$, where $J_i$ is the integral of squared jerk. 
  This acts as a reliable proxy for trajectory aggressiveness and smoothness costs.

  \item Residual Trend ($\tau_{\text{trend}}$): 
  Defined as $\tanh(k_{\text{slope}})$, where $k_{\text{slope}}$ is the slope of a linear regression fitted to the segment's recent log-residual history. 
  A negative value indicates a decreasing residual (active convergence), while a value near zero signals stagnation.

  \item Segment Duration ($T_i$): 
  The allocated duration of the $i$-th segment.
  It guards against numerical ill-conditioning from over-splitting and serves as the baseline for the time inflation action.

  \item Boundary Bias ($\delta_i$): 
  Defined as $\delta_i = \tanh(\epsilon_{i}^{\text{right}} - \epsilon_{i}^{\text{left}})$. 
  A positive value indicates higher residuals at the right boundary, guiding the policy to shift the split point leftward to allocate more degrees of freedom to the constrained side.
\end{itemize}



\textbf{Global State ($\mathbf{s}^{\text{glo}}$):}
To capture the overall solver status independent of any specific segment, we construct a global context shared across all agents:
\begin{equation}
\mathbf{s}^{\text{glo}} = \Big[~ \|\boldsymbol{\epsilon}\|_\infty,~ \|\mathbf{y}\|_\infty,~ \|\mathbf{\bar{J}}\|_\infty,~ \bar{\boldsymbol{\epsilon}},~ N/N_{\max} ~\Big]^\top.
\end{equation}
Obtained via a ``Max-Pooling'' operation across all segments, the first three metrics capture the current worst-case convergence bottlenecks.
The fourth metric, $\bar{\boldsymbol{\epsilon}}$, provides the mean residual as a reference baseline.
Rather than a hard structural cap, the ratio $N/N_{\max}$ serves as a \textit{soft budgetary signal}.


\subsubsection{Action Space}
To decouple the high-level splitting decision from the low-level spatio-temporal adjustment, we design a continuous action space. 
For each agent $i$, the shared policy $\pi_\theta(\mathbf{s}_i)$ outputs an action vector $\mathbf{a}_i = [a_i^{\text{gate}}, a_i^{\text{ratio}}, a_i^{\text{bias}}, a_i^{\text{inf}}]^\top \in [-1, 1]^4$. 
Each component controls a distinct aspect of the re-splitting process:
\begin{itemize}
\item Splitting Propensity ($a_i^{\text{gate}}$): Encodes the agent's urgency to trigger a split.
A split is triggered only if $a_i^{\text{gate}} > \delta_{\text{gate}}$.
Moreover, the raw magnitude serves as a ``bid'' during the election process (Sec. \ref{Sec3D:vote}), prioritizing the most stagnating agents at each step.
\item Spatial Split Ratio ($a_i^{\text{ratio}}$):
Determines the spatial split location.
To avoid extremely short segments that lead to ill-conditioned subproblems, the raw output is linearly mapped to a safe range as $0.4 a_i^{\text{ratio}} + 0.5 \in [0.1, 0.9]$.

\item Temporal Bias ($a_i^{\text{bias}}$):
Decouples time allocation from the spatial split geometry. 
With a temporal bias of $0.2 a_i^{\text{bias}}$ added to the mapped spatial ratio, the actual time split ratio is computed as $\text{clip}(0.4 a_i^{\text{ratio}} + 0.5 + 0.2 a_i^{\text{bias}},\, 0.1, 0.9)$. 
This allows the agent to assign more reasonable relative durations to the two new segments, rather than strictly mirroring the spatial ratio.

\item Duration Inflation ($a_i^{\text{inf}}$):
Addresses the problem that strictly dividing the original time allocation may leave the new segments with insufficient durations.
The resulting factor $\eta = \max(0, 0.3 a_i^{\text{inf}})$ expands the total duration by up to 30\%.
This additional time ensures the split segments have sufficient temporal flexibility to satisfy dynamic constraints.
\end{itemize}

\subsubsection{Reward Function}
To align the agents' learning goals with the ADMM solver's convergence behavior, we design a composite reward function. 
Rather than relying on sparse terminal signals, we decompose the reward into a System-Level Convergence Reward ($r_{\text{sys}}$) and an Action-Level Regularization ($r_{\text{act}}$):
\begin{equation}
r_{i,t} = r_{\text{sys}} + \mathbb{I}_{\text{split}} \cdot r_{\text{act}},
\end{equation}
where $\mathbb{I}_{\text{split}}$ is the indicator function triggering the regularization terms exclusively for the agent executing the split.

\textbf{System-Level Convergence Reward ($r_{\text{sys}}$):} This component evaluates the overall optimization progress at each step. 
It consists of three parts:
\begin{itemize}
  \item {Progress Reward ($r_{\text{prog}}$):} Rewards the per-step reduction of the worst-case residual in log-scale:
  \begin{equation}
    r_{\text{prog}} = \lambda_1 \left( \log_{10} \|\boldsymbol{\epsilon}\|_\infty^{(t-1)} - \log_{10} \|\boldsymbol{\epsilon}\|_\infty^{(t)} \right).
  \end{equation}
  \item {Time Penalty ($r_{\text{step}}$):} A constant penalty $r_{\text{step}} = -\lambda_2$ is applied per step to encourage the policy to minimize the total iteration count.
  \item {Terminal Outcome ($r_{\text{term}}$):} Upon termination, failed cases receive a penalty $-R_{\text{fail}}$. 
  For successful convergence, the reward is defined as:
    \begin{equation}
    r_{\text{term}} = 
    \begin{cases} 
    0.5 R_{\text{conv}}, & \!\! N \ge N_{\text{max}} \\
    R_{\text{conv}} + \lambda_3 (K_{\text{base}} \!-\! K_{\text{ours}}), & \!\! N < N_{\text{max}}
    \end{cases}
    \end{equation}
  where $R_{\text{conv}}$ is the base success reward, halved when the segment count reaches the soft budget $N_{\text{max}}$ to discourage excessive splitting.
  The second condition introduces a bonus proportional to the iteration saving $(K_{\text{base}} - K_{\text{ours}})$ over the baseline \cite{TOP}.
\end{itemize}

\textbf{Action-Level Regularization ($r_{\text{act}}$):} 
To regularize the splitting behavior, we impose the following terms on the splitting action:
\begin{itemize}
  \item {Splitting Penalty ($r_{\text{split}}$):} 
  A constant penalty $r_{\text{split}} = -\lambda_4$ is applied to each split action to discourage unnecessary structural changes that incur computational overhead. 
  \item {Energy Balance ($r_{\text{bal}}$):} A high-quality split should distribute the dynamic load evenly.
  We penalize the jerk density imbalance between the two resulting sub-segments:
  \begin{equation}
        r_{\text{bal}} = -\lambda_5 \left| \bar{J}_{\text{left}} - \bar{J}_{\text{right}} \right|.
  \end{equation}
  \item {Inflation Penalty ($r_{\text{inf}}$):} While temporal inflation improves feasibility, excessive dilation yields overly conservative trajectories. 
  We penalize the excess duration beyond the original allocation: $r_{\text{inf}} = -\lambda_6 \, \eta.$
  \item {Consistency Guidance ($r_{\text{guide}}$):} 
  To accelerate early-stage training, we introduce a shaping reward that encourages the split point to shift away from the side with higher residuals.
  Recall that a positive Boundary Bias ($\delta_i > 0$) indicates higher residuals at the right boundary.
  Accordingly, the policy should place the split point leftward (i.e., $a_i^{\text{ratio}} < 0$) to assign more degrees of freedom to the right side.
  The guidance is defined as:

  \begin{equation}
        r_{\text{guide}} = -\lambda_7 \left[ -\delta_i (a_i^{\text{ratio}} + a_i^{\text{bias}}) \right] \cdot \zeta(k).
  \end{equation}
  Here, $\zeta(k)$ is a decay factor that gradually reduces the influence of this shaping term as training progresses.
\end{itemize}

\begin{figure*}[!ht]
  \centering
  \includegraphics[width=0.95\textwidth]{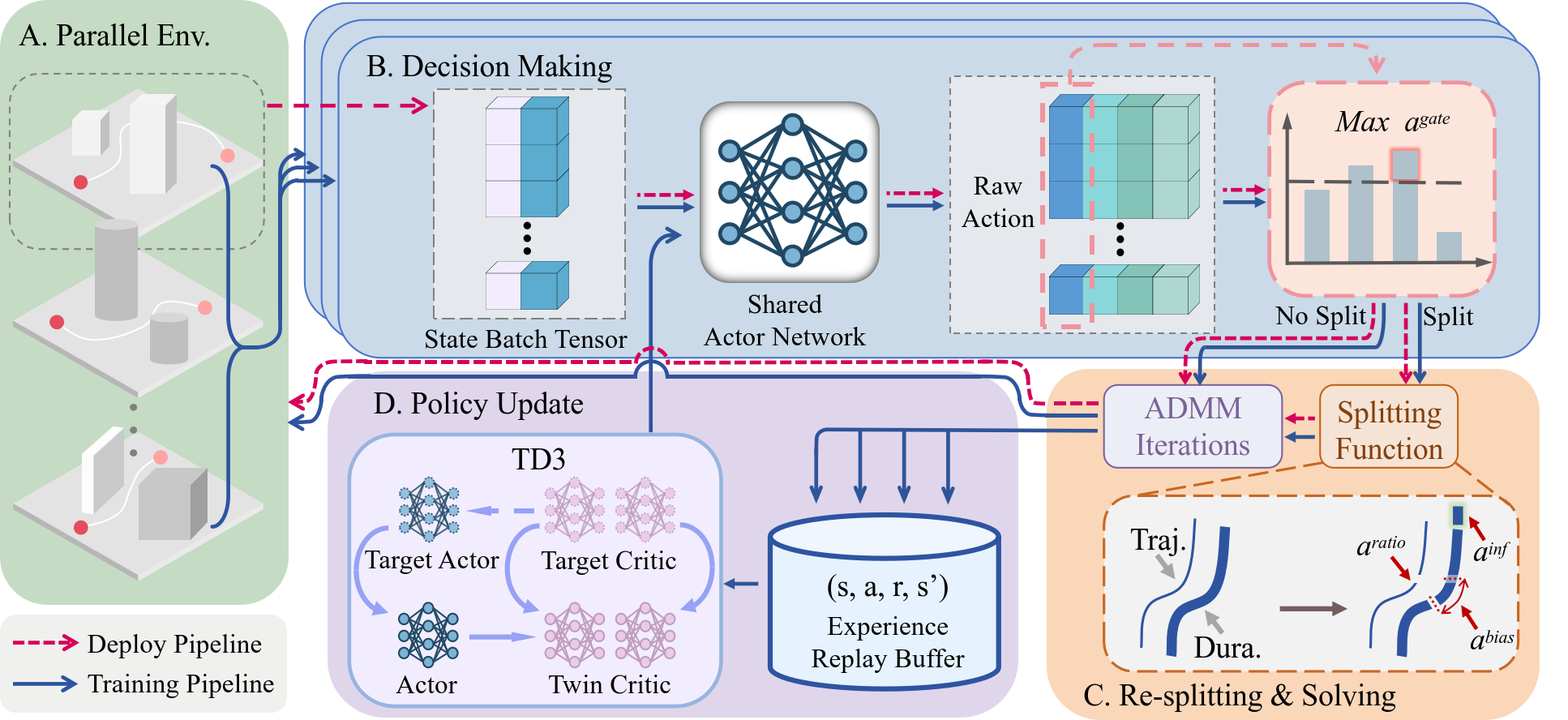}
  \caption{System architecture.
  The system consists of four modules: (A) Parallel Environments, (B) Decision Making via Shared Policy Inference and Confidence-Based Election, (C) Trajectory Re-splitting and ADMM Solving, and (D) Centralized TD3 Policy Update.
  During training (solid blue lines), A produces parallel rollouts that flow through B and C, with transitions stored in the replay buffer for off-policy updates in D.
  Updated states are fed back to A to form a closed loop.
  During deployment (red dashed lines), a single environment uses the pre-trained actor in B and solver in C, bypassing D entirely.
}
  \label{fig:pipeline}
  \vspace{-0.4cm}
\end{figure*}

\vspace{-0.3cm}
\subsection{Policy Design}
\label{Sec3C:policy design} 
\subsubsection{Algorithm Selection}
Since each environment step requires executing multiple ADMM iterations, generating training samples is computationally expensive.
This necessitates an off-policy algorithm that can maximize data reuse. 
We therefore adopt the {Twin Delayed Deep Deterministic Policy Gradient (TD3)} algorithm \cite{td3}, whose replay buffer mechanism provides superior sample efficiency. 
Furthermore, its twin-critic design stabilizes training despite the high variance inherent in the optimization landscape.

\subsubsection{Network Architecture}
To balance inference speed with representational capacity, we devise a specialized Actor-Critic architecture:
\begin{itemize}
  \item \textbf{Actor Network ($\pi_\theta$):} 
  The Actor maps the state $\mathbf{s}_i$ to the action space. 
  It features a shared encoder (two fully connected layers, LayerNorm, and ReLU) to extract high-level features. 
  The output layer branches into four independent specialized heads. 
  This multi-head design decouples the splitting decision ($a_i^{\text{gate}}$) from the spatio-temporal adjustments ($a_i^{\text{ratio}}, a_i^{\text{bias}}, a_i^{\text{inf}}$), preventing gradient interference between these two objectives.
  We apply a Tanh activation to bound all four outputs to $[-1, 1]$.
  \item \textbf{Critic Network ($Q_\phi$):} 
  The Critic takes the concatenated state and action $(\mathbf{s}_i, \mathbf{a}_i)$ as input and estimates the Q-value through two fully connected layers (with LayerNorm and ReLU), followed by a linear projection to a scalar output. 
  Following the TD3 paradigm, we employ twin critics ($Q_{\phi_1}, Q_{\phi_2}$) with identical architecture but independent parameters, and take the minimum of their outputs as the target to mitigate value overestimation bias.

\end{itemize}

\subsection{Confidence-Based Election Mechanism}
\label{Sec3D:vote}
Simultaneous structural changes across multiple segments can destabilize the optimization process and hinder convergence.
To prevent this, we propose a Confidence-Based Election mechanism that enforces single-winner updates.
Specifically, the Actor's splitting propensity $a_i^{\text{gate}}$ also serves as the election bid, avoiding the need for an additional coordination module.
At each step, only the agent with the highest propensity above the threshold $\delta_{\text{gate}}$ is selected to execute the split:
\begin{equation}
i^* = \arg \max_{i} \{ a_i^{\text{gate}} \mid a_i^{\text{gate}} > \delta_{\text{gate}} \}
\end{equation}
This ensures that at most one structural change occurs per step, targeting the most stagnating segment.

\vspace{-0.2cm}
\subsection{Training and Deployment Pipeline}
\label{Sec3E:train}
As illustrated in Fig. \ref{fig:pipeline}, ATRS operates as a closed-loop pipeline across four modules.

During training (solid blue lines), multiple parallel environments (Module A) generate diverse rollouts simultaneously.
Observations are concatenated into a State Batch Tensor and fed to the Shared Actor Network (Module B), which outputs raw actions for all agents.
The Confidence-Based Election selects the winning agent, whose action is passed to Module C, where the Splitting Function applies the re-splitting and the ADMM solver iterates to update the trajectory.
Updated states are fed back to Module A, completing one interaction cycle.
The collected transitions $(s, a, r, s')$ are pushed into a centralized Experience Replay Buffer, from which Module D samples mini-batches to update the Actor and Twin Critic networks.

During deployment (red dashed lines), the system operates on a single environment.
The same forward path through Modules A, B, and C is preserved, but Module D is bypassed entirely, and the pre-trained Actor directly guides the ADMM solver for real-time planning.


\begin{figure}[!b]
  \centering
  \includegraphics[width=0.476\textwidth]{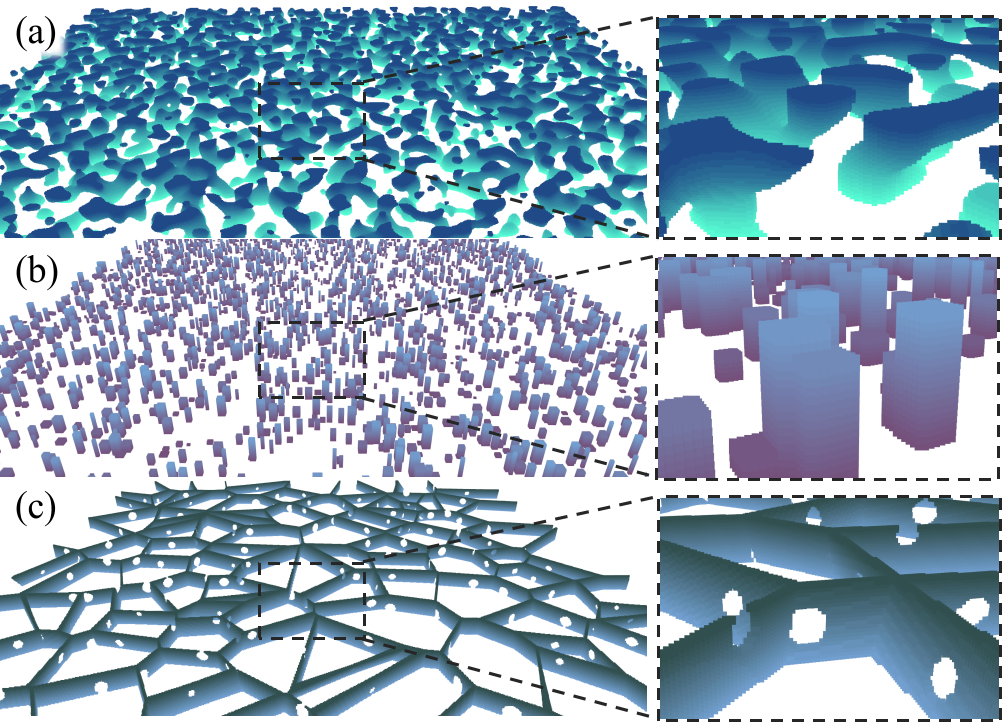}
  \caption{Overview of the benchmark environments. 
  (a) {Map A:} Sparse obstacle distribution generated by 3D Perlin noise (used for training).
  (b) {Map B:} A highly cluttered environment containing 2,500 random cubic obstacles.
  (c) {Map C:} A structured environment with 120 irregular polygonal cells connected by narrow passages.
  }
  \label{fig:maps}
\end{figure}

\section{Simulations}
\label{Sec: Simulations}
\subsection{Implementation Details}
\label{Sec: Implementation}

%
We implement the complete ATRS framework in {LibTorch} (PyTorch C++ frontend). 
This enables native integration with the ADMM solver and the broader C++ planning pipeline without cross-language overhead.
The resulting unified C++ pipeline achieves millisecond-level inference latency, suitable for onboard deployment.
To maximize training throughput, we decouple data generation from policy learning.
A custom \texttt{ProblemGenerator} pre-computes diverse problem instances, including flight corridors and initial paths.
These instances are serialized into JSON format for offline storage.
During training, we adopt a {hybrid parallelization scheme}: {OpenMP} distributes concurrent environment interactions, while {Intel TBB} accelerates the internal ADMM optimization.
Meanwhile, neural network inference requests are batched to maximize GPU utilization.
All simulations run on a workstation equipped with an Intel i5-12600KF CPU and an NVIDIA RTX 4070 Ti GPU. 
To ensure reproducibility, we fix random seeds across all libraries. 
Using the hyperparameters detailed in Table \ref{tab:params}, the policy is trained via TD3 with an action noise annealing schedule.
The policy converges within 10,000 episodes in approximately 10 minutes.
\begin{table}[h]
\centering
\caption{Hyperparameters for TD3 Training}
\label{tab:params}
\resizebox{\columnwidth}{!}{%
\begin{tabular}{lclc}
\toprule
\textbf{Hyperparameter} & \textbf{Value} & \textbf{Hyperparameter} & \textbf{Value} \\ 
\midrule
Actor Learning Rate & $4 \times 10^{-4}$ & Batch Size & 512 \\
Critic Learning Rate & $4 \times 10^{-3}$ & Buffer Capacity & $4 \times 10^4$ \\
Discount Factor $\gamma$ & 0.99 & Hidden Dim & 256 \\
Soft Update $\tau$ & 0.005 & Threshold $\delta_{\text{gate}}$ & $0.3$ \\
Exploration Noise $\sigma$ & $0.1 \to 0.01$ & Target Policy Noise & 0.2 \\
Policy Delay Freq & 4  & Target Noise Clip & 0.5 \\

\bottomrule
\end{tabular}%
}
\vspace{-0.4cm}
\end{table}

\subsection{Benchmark Comparisons}
\label{Sec: Benchmark}
To validate the effectiveness and generalization of {ATRS}, we conduct experiments across varying obstacle densities and problem scales.

\subsubsection{Experimental Setup \& Baselines}
The experimental settings, baselines, and evaluation metrics are detailed below.

\textbf{Environment Settings:}
Simulations are conducted in a $200\,\text{m} \times 200\,\text{m} \times 5\,\text{m}$ workspace, with start and goal configurations randomly sampled from collision-free regions.
The environment is generated using 3D Perlin noise (Fig.~\ref{fig:maps}(a)), which produces spatially correlated obstacle clusters.
To evaluate robustness under distribution shifts, we define the occupancy ratio $\rho$ as the fraction of obstacle voxels in the 3D grid.
We define three density levels: \textit{Sparse} ($\rho=0.2$, used for training), \textit{Medium} ($\rho=0.3$), and \textit{Dense} ($\rho=0.4$).
Furthermore, problem instances are categorized by trajectory length $L$ into three scales: \textit{Short} ($L \le 50\,\text{m}$), \textit{Medium} ($50 < L \le 150\,\text{m}$), and \textit{Long} ($L > 150\,\text{m}$).

\textbf{Baselines:} We compare {ATRS} with two baseline methods:
\begin{itemize}
\item[a)] {TOP (Fixed-Structure) \cite{TOP}:} The state-of-the-art parallel trajectory optimizer based on CADMM.
It fixes the number of segments after initialization, serving as a baseline for fixed-structure approaches.
\item[b)] {Heuristic (Rule-Based):} Since no prior method addresses adaptive re-splitting, we construct a rule-based baseline to isolate the gains from learned adaptation.
After initialization, it deterministically selects the segment with the highest residual and splits it with a fixed spatial and temporal ratio of $0.5$.
\end{itemize}
To ensure a fair comparison, the reported total time for ATRS includes both the ADMM solver runtime and the neural network inference latency.

\textbf{Metrics:} Performance is quantified using four key metrics:
\begin{itemize}
\item {Iterations:} Average ADMM iterations to convergence.
\item {Total Time:} The wall-clock time including both ADMM solver computation and neural network inference.
\item {Energy Cost:} A measure of trajectory smoothness given by the integral of squared jerk, $J = \int \| \mathbf{j}(t) \|^2 \df{t}$.
\item {Success Rate:} The percentage of trials that successfully converge to a feasible solution within 2,000 iterations, following the stopping criteria defined in \cite{Boyd}.
\end{itemize}

\begin{table*}[t]
  \vspace*{2pt}
  \centering
  \caption{Benchmark comparison under varying environmental densities ($\rho$) and problem scales.\\
  Iter.: Iterations; Time: Total Time; Cost: Energy Cost; SR: Success Rate.}
  \label{tab:benchmark_results}
  \renewcommand{\arraystretch}{1.2}
  \setlength{\tabcolsep}{3.7pt}
  \begin{tabular}{c c cccc cccc cccc}
    \toprule[1.5pt]
    \multirow{2}{*}{\makecell{Env.\\Density}} & \multirow{2}{*}{Method} & \multicolumn{4}{c}{Short ($<50m$)} & \multicolumn{4}{c}{Medium ($50 \sim 150m$)} & \multicolumn{4}{c}{Long ($>150m$)}  \\
    \cmidrule(lr){3-6} \cmidrule(lr){7-10} \cmidrule(l){11-14}
    & & Iter. & Time (ms) & Cost & SR (\%) & Iter. & Time (ms) & Cost & SR (\%) & Iter. & Time (ms) & Cost & SR (\%) \\
    \midrule
    
    \multirow{3}{*}{\makecell{Sparse\\($\rho=0.2$)}} 
     & Heuristic       & 510.2 & 34.2 & 54.5 & 99 & 469.2 & 45.8 & 54.0 & 100 & 578.1 & 85.1 & 83.2 & 98 \\
     & TOP  & 412.9 & 25.6 & 26.5 & 97 & 432.7 & 42.3 & 34.1 & 98 & 441.9 & 64.7 & 56.6 & 99 \\
     & \cellcolor{gray!15}\textbf{ATRS} & \cellcolor{gray!15}\textbf{305.4} & \cellcolor{gray!15}\textbf{20.7} & \cellcolor{gray!15}\textbf{23.5} & \cellcolor{gray!15}\textbf{100} & \cellcolor{gray!15}\textbf{356.1} & \cellcolor{gray!15}\textbf{37.9} & \cellcolor{gray!15}\textbf{32.1} & \cellcolor{gray!15}\textbf{100} & \cellcolor{gray!15}\textbf{392.2} & \cellcolor{gray!15}\textbf{61.5} & \cellcolor{gray!15}\textbf{53.6} & \cellcolor{gray!15}\textbf{100} \\
    \midrule
        
    \multirow{3}{*}{\makecell{Medium\\($\rho=0.3$)}}
     & Heuristic       & 459.2 & 32.2 & 51.6 & 99 & 539.3 & 56.1 & 71.0 & 98 & 520.1 & 81.6 & 79.6 & 100 \\
     & TOP  & 415.2 & 26.7 & 25.9 & 99 & 446.4 & 46.2 & 45.6 & 98 & 420.7 & 66.8 & 57.6 & 99 \\
     & \cellcolor{gray!15}\textbf{ATRS} & \cellcolor{gray!15}\textbf{307.9} & \cellcolor{gray!15}\textbf{22.2} & \cellcolor{gray!15}\textbf{23.8} & \cellcolor{gray!15}\textbf{99} & \cellcolor{gray!15}\textbf{381.3} & \cellcolor{gray!15}\textbf{42.7} & \cellcolor{gray!15}\textbf{42.2} & \cellcolor{gray!15}\textbf{100} & \cellcolor{gray!15}\textbf{390.7} & \cellcolor{gray!15}\textbf{65.4} & \cellcolor{gray!15}\textbf{55.6} & \cellcolor{gray!15}\textbf{100} \\
    \midrule
        
    \multirow{3}{*}{\makecell{Dense\\($\rho=0.4$)}} 
     & Heuristic       & 545.1 & 38.6 & 57.2 & 96 & 689.9 & 72.0 & 75.7 & 98 & 816.7 & 123.1 & 77.5 & 96 \\  
     & TOP  & 442.8 & 29.2 & 28.0 & 94 & 474.8 & 50.4 & 42.6 & 98 & 454.4 & 70.8 & 42.7 & 100 \\
     & \cellcolor{gray!15}\textbf{ATRS} & \cellcolor{gray!15}\textbf{343.0} & \cellcolor{gray!15}\textbf{24.4} & \cellcolor{gray!15}\textbf{25.2} & \cellcolor{gray!15}\textbf{97} & \cellcolor{gray!15}\textbf{438.5} & \cellcolor{gray!15}\textbf{50.2} & \cellcolor{gray!15}\textbf{41.1} & \cellcolor{gray!15}\textbf{99} & \cellcolor{gray!15}\textbf{431.9} & \cellcolor{gray!15}\textbf{70.8} & \cellcolor{gray!15}\textbf{41.9} & \cellcolor{gray!15}\textbf{100} \\
    \bottomrule[1.5pt]
  \end{tabular}%
  \vspace{-0.4cm}
\end{table*}

\subsubsection{Analysis of Results}
Table~\ref{tab:benchmark_results} presents the quantitative comparison on Map A.
Although the policy was trained only under \textit{Sparse} density, it is evaluated directly on unseen \textit{Medium} and \textit{Dense} levels without retraining.
All experiments are repeated 100 times per setting.
The results demonstrate that {ATRS} outperforms both baselines across all problem scales and obstacle densities.

Although the Heuristic baseline also performs re-splitting, it consistently yields the worst results.
Its strategy leads to substantially higher energy cost and more iterations than ATRS across all settings.
This highlights that \textit{where} and \textit{how} to re-split matter far more than simply re-splitting with a handcrafted rule.

Compared to the fixed-structure TOP solver, ATRS achieves consistent improvements.
In the best case (\textit{Short}/\textit{Sparse}), ATRS reduces the number of iterations by 26.0\%, yielding a 19.1\% reduction in wall-clock time after accounting for inference latency.
As density increases from \textit{Sparse} to \textit{Dense}, the iteration advantage narrows from 26.0\% to 22.5\% (\textit{Short} scale), which is expected since denser obstacles shrink the feasible space and the policy must generalize to unseen distributions.
At larger scales, the number of segments exceeds the available parallel threads. 
Additional splits can no longer run fully in parallel and instead incur thread scheduling overhead, narrowing the wall-clock advantage.
Nevertheless, ATRS achieves fewer iterations, lower energy cost, and higher success rates across all nine settings, including unseen density conditions.
\begin{figure}[th]
    \centering
    \includegraphics[width=0.95\linewidth]{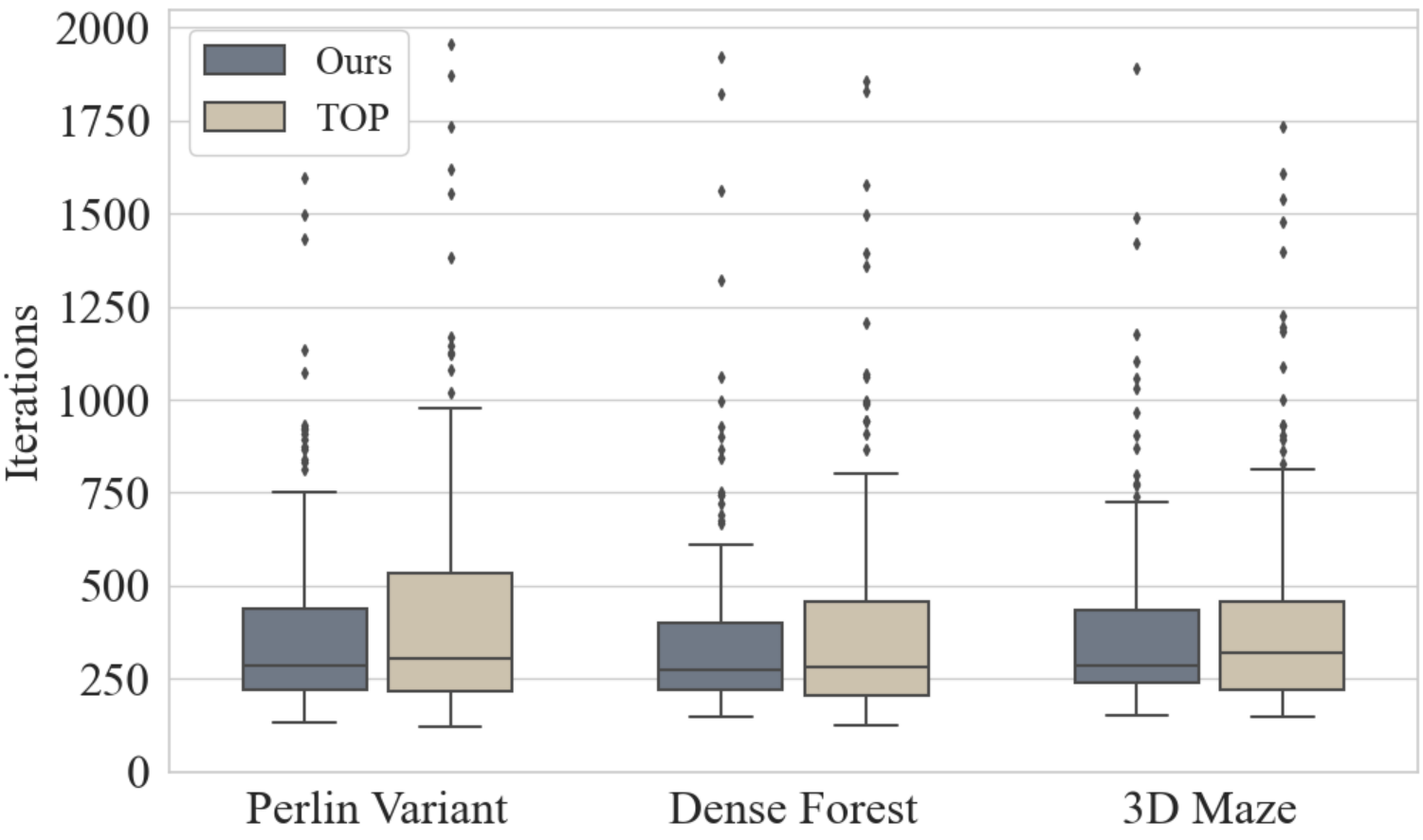}
    \caption{Iteration distribution across unseen environments. 
    ATRS (gray) exhibits a lower median and less variance than TOP (brown) in all three scenarios.}
    \label{fig:boxplot_iterations}
    \vspace{-15pt}
\end{figure}

\subsubsection{Generalization Verification}
To further validate zero-shot generalization, we evaluate ATRS on three unseen environments, including an in-distribution variant and two structurally distinct layouts (Fig.~\ref{fig:maps}(b)-(c)):
\begin{itemize}
    \item {Map A-2 (Perlin Variant):} A variant generated via Perlin noise with an altered random seed, following the same distribution as the training map (Map A).
    \item {Map B (Dense Forest):} A highly cluttered environment populated with 2,500 discrete cubic obstacles.
    \item {Map C (3D Maze):} A structured environment with 120 irregular polygonal cells connected by narrow passages.
\end{itemize}

Each scenario is evaluated over 100 independent trials.
As shown in Fig.~\ref{fig:boxplot_iterations}, ATRS achieves a consistently lower median iteration count than TOP across all three unseen environments.
More importantly, ATRS exhibits substantially reduced variance and fewer long-tail outliers.
Such outliers correspond to segments that stagnate under tight local constraints, hindering the global CADMM consensus.
ATRS alleviates this by adaptively re-splitting such segments, markedly reducing the frequency of extreme cases.
Even in the 3D Maze, whose narrow passages and structured topology differ fundamentally from the training environment, ATRS maintains a tighter iteration distribution.

This cross-scenario robustness stems from our geometry-agnostic state design.
Since the design relies solely on the solver's internal states rather than geometric features, the policy captures the invariant numerical dynamics of the optimization process.
Consequently, it generalizes to environments with different obstacle structures without retraining.

\begin{figure}[!b]
  \vspace{-10pt}
    \centering
    \includegraphics[width=0.95\linewidth]{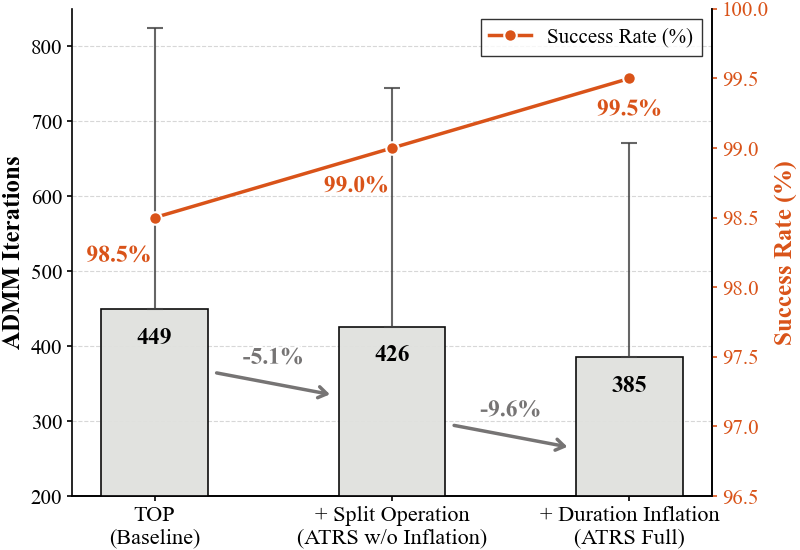}
    \caption{Ablation study on Map A. 
    Left axis shows ADMM iterations (bars); right axis shows success rate (line).
    }
    \label{fig:ablation}
\end{figure}

\vspace{-10pt}
\subsection{Ablation Study}
\label{Sec: Ablation}
To isolate the contributions of re-splitting and duration inflation, we conduct an ablation study on Map A with 200 independent trials.
Specifically, we evaluate \textbf{ATRS w/o Inflation}, an ablated variant where the duration inflation action is explicitly masked. 
As illustrated in Fig.~\ref{fig:ablation}, even without duration inflation, this variant reduces ADMM iterations by 5.1\% (426 vs. 449) and improves the success rate from 98.5\% to 99.0\% compared to TOP. 
This confirms that structural re-splitting alone already provides measurable gains.
Restoring the duration inflation action in \textbf{ATRS Full} yields a further 9.6\% iteration reduction (385 vs. 426) and raises the success rate to {99.5\%}.
This progressive improvement confirms that the two actions are synergistic.
Re-splitting provides the structural foundation by injecting additional degrees of freedom, while duration inflation ensures sufficient temporal allocation for the newly created segments to converge.



\vspace{-10pt}
\subsection{Real-World Experiments}
\label{Sec: Real-world}

We validate ATRS on a customized quadrotor equipped with an NVIDIA Orin NX and a Livox MID360 LiDAR.
The onboard computer simultaneously runs FAST-LIO2~\cite{FastLio2} for state estimation and mapping, along with planning and control.

\begin{figure}[t]
    \centering
    \includegraphics[width=0.95\linewidth]{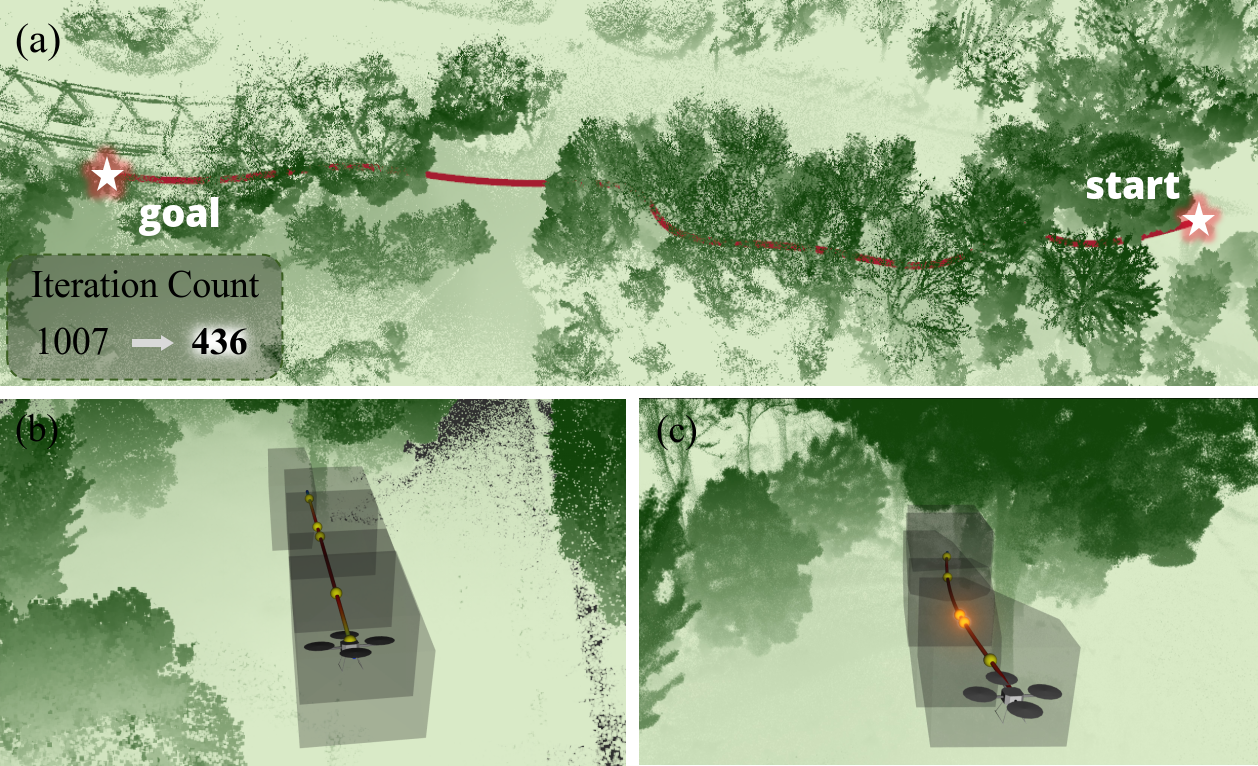}
    \caption{Real-world experiments.
(a) Global trajectory (85\,m, 20 segments) planned on a pre-built point-cloud map. ATRS reduces iterations from 1007 to 436.
(b) Online replanning in a sparse region; no split occurs.
(c) Online replanning near dense obstacles; orange dots mark the boundaries of the adaptively inserted segment.}
    \label{fig:real_world}
    \vspace{-10pt}
\end{figure}

\textbf{Offline global planning.}
To evaluate the parallel optimization capability on a large-scale problem, we construct a point-cloud map of the test site and plan an 85\,m trajectory offline (Fig.~\ref{fig:real_world}(a)).
ATRS reduces the iteration count from 1007 to {436}, a {56.7\%} reduction compared to the fixed-structure baseline.

\textbf{Online local planning.}
We further deploy ATRS as an online local planner in unknown environments using the SUPER~\cite{super} framework.
As shown in Fig.~\ref{fig:real_world}(b)-(c), the policy adapts its behavior to local optimization difficulty: in sparse regions the trajectory converges without splitting, while near dense obstacles the agent triggers splits.
Across 80 consecutive replanning cycles, the system maintains an average latency of 35\,ms per cycle (including neural network inference) and sustains a maximum speed of 3.2\,m/s.
These results confirm that ATRS is lightweight enough for onboard deployment and that the policy trained in simulation transfers to the real world without fine-tuning.



\vspace{-0.2cm}
\section{Conclusion}
In this paper, we presented {ATRS}, which is based on parallel ADMM and embeds a shared DRL policy to re-split stagnating trajectory segments online.
The re-splitting problem is formulated as a MASP-MDP in which all segments act as homogeneous agents sharing a unified policy network.
Parameter sharing renders the policy size-invariant, allowing it to scale to trajectories of arbitrary length without retraining.
Since the state representation depends only on the solver's internal states rather than geometric features of the environment, the policy transfers zero-shot to unseen scenarios.
A Confidence-Based Election mechanism safeguards solver stability by admitting only the most stagnating segment for re-splitting at each step.
Across diverse simulation benchmarks, ATRS achieved up to 26.0\% fewer iterations and 19.1\% shorter wall-clock time than fixed-structure baselines.
Onboard deployment on a resource-constrained quadrotor delivered a 56.7\% iteration reduction in offline global planning and sustained local replanning at 35\,ms per cycle, with no sim-to-real degradation.

\bibliographystyle{IEEEtran}
\bibliography{paper_resub}

\begin{thebibliography}{10}
\providecommand{\url}[1]{#1}
\csname url@samestyle\endcsname
\providecommand{\newblock}{\relax}
\providecommand{\bibinfo}[2]{#2}
\providecommand{\BIBentrySTDinterwordspacing}{\spaceskip=0pt\relax}
\providecommand{\BIBentryALTinterwordstretchfactor}{4}
\providecommand{\BIBentryALTinterwordspacing}{\spaceskip=\fontdimen2\font plus
\BIBentryALTinterwordstretchfactor\fontdimen3\font minus \fontdimen4\font\relax}
\providecommand{\BIBforeignlanguage}[2]{{%
\expandafter\ifx\csname l@#1\endcsname\relax
\typeout{** WARNING: IEEEtran.bst: No hyphenation pattern has been}%
\typeout{** loaded for the language `#1'. Using the pattern for}%
\typeout{** the default language instead.}%
\else
\language=\csname l@#1\endcsname
\fi
#2}}
\providecommand{\BIBdecl}{\relax}
\BIBdecl

\bibitem{environmental_monitoring}
A.~Fascista, ``Toward integrated large-scale environmental monitoring using wsn/uav/crowdsensing: A review of applications, signal processing, and future perspectives,'' \emph{Sensors}, vol.~22, no.~5, p. 1824, 2022.

\bibitem{dong2025multi}
H.~Dong, X.~Ma, and S.~Zhang, ``Multi-flight path planning for a single agricultural drone in a regular farmland area,'' \emph{Sustainability}, vol.~17, no.~6, p. 2433, 2025.

\bibitem{GPOPS-II}
M.~A. Patterson and A.~V. Rao, ``Gpops-ii: A matlab software for solving multiple-phase optimal control problems using hp-adaptive gaussian quadrature collocation methods and sparse nonlinear programming,'' \emph{ACM Transactions on Mathematical Software (TOMS)}, vol.~41, no.~1, pp. 1--37, 2014.

\bibitem{trajOpt}
J.~Schulman, J.~Ho, A.~X. Lee, I.~Awwal, H.~Bradlow, and P.~Abbeel, ``Finding locally optimal, collision-free trajectories with sequential convex optimization.'' in \emph{Robotics: Science and Systems}, vol.~9, no.~1.\hskip 1em plus 0.5em minus 0.4em\relax Berlin, Germany, 2013, pp. 1--10.

\bibitem{Gcopter}
Z.~Wang, X.~Zhou, C.~Xu, and F.~Gao, ``Geometrically constrained trajectory optimization for multicopters,'' \emph{IEEE Transactions on Robotics}, vol.~38, no.~5, pp. 3259--3278, 2022.

\bibitem{trajectorySplitting}
C.~Wang, J.~Bingham, and M.~Tomizuka, ``Trajectory splitting: A distributed formulation for collision avoiding trajectory optimization,'' in \emph{2021 IEEE/RSJ International Conference on Intelligent Robots and Systems (IROS)}.\hskip 1em plus 0.5em minus 0.4em\relax IEEE, 2021, pp. 8113--8120.

\bibitem{TOP}
J.~Yu, N.~Chen, G.~Liu, C.~Xu, F.~Gao, and Y.~Cao, ``Top: Trajectory optimization via parallel optimization towards constant time complexity,'' \emph{IEEE Robotics and Automation Letters}, 2025.

\bibitem{D*A*}
T.~H. Cormen, C.~E. Leiserson, R.~L. Rivest, and C.~Stein, \emph{Introduction to algorithms}.\hskip 1em plus 0.5em minus 0.4em\relax MIT press, 2022.

\bibitem{sambharya2024learning}
R.~Sambharya, G.~Hall, B.~Amos, and B.~Stellato, ``Learning to warm-start fixed-point optimization algorithms,'' \emph{Journal of Machine Learning Research}, vol.~25, no. 166, pp. 1--46, 2024.

\bibitem{rlqp}
J.~Ichnowski, P.~Jain, B.~Stellato, G.~Banjac, M.~Luo, F.~Borrelli, J.~E. Gonzalez, I.~Stoica, and K.~Goldberg, ``Accelerating quadratic optimization with reinforcement learning,'' \emph{Advances in Neural Information Processing Systems}, vol.~34, pp. 21\,043--21\,055, 2021.

\bibitem{kumarMinjerk}
D.~Mellinger and V.~Kumar, ``Minimum snap trajectory generation and control for quadrotors,'' in \emph{IEEE International Conference on Robotics and Automation (ICRA)}.\hskip 1em plus 0.5em minus 0.4em\relax IEEE, 2011, pp. 2520--2525.

\bibitem{navrl}
Z.~Xu, X.~Han, H.~Shen, H.~Jin, and K.~Shimada, ``Navrl: Learning safe flight in dynamic environments,'' \emph{IEEE Robotics and Automation Letters}, 2025.

\bibitem{neo-planner}
Y.~Chen, J.~Li, W.~Qin, Y.~Hua, X.~Dong, and Q.~Li, ``Learning to initialize trajectory optimization for vision-based autonomous flight in unknown environments,'' in \emph{2025 IEEE/RSJ International Conference on Intelligent Robots and Systems (IROS)}, 2025, pp. 9525--9532.

\bibitem{wu2024deep}
Y.~Wu, X.~Sun, I.~Spasojevic, and V.~Kumar, ``Deep learning for optimization of trajectories for quadrotors,'' \emph{IEEE Robotics and Automation Letters}, vol.~9, no.~3, pp. 2479--2486, 2024.

\bibitem{huang2020one}
W.~Huang, I.~Mordatch, and D.~Pathak, ``One policy to control them all: Shared modular policies for agent-agnostic control,'' in \emph{International Conference on Machine Learning}.\hskip 1em plus 0.5em minus 0.4em\relax PMLR, 2020, pp. 4455--4464.

\bibitem{lowe2017multi}
R.~Lowe, Y.~I. Wu, A.~Tamar, J.~Harb, O.~Pieter~Abbeel, and I.~Mordatch, ``Multi-agent actor-critic for mixed cooperative-competitive environments,'' \emph{Advances in Neural Information Processing Systems}, vol.~30, 2017.

\bibitem{littman1994markov}
M.~L. Littman, ``Markov games as a framework for multi-agent reinforcement learning,'' in \emph{Machine learning proceedings 1994}.\hskip 1em plus 0.5em minus 0.4em\relax Elsevier, 1994, pp. 157--163.

\bibitem{td3}
S.~Fujimoto, H.~Hoof, and D.~Meger, ``Addressing function approximation error in actor-critic methods,'' in \emph{International Conference on Machine Learning}.\hskip 1em plus 0.5em minus 0.4em\relax PMLR, 2018, pp. 1587--1596.

\bibitem{Boyd}
S.~Boyd, N.~Parikh, E.~Chu, B.~Peleato, J.~Eckstein \emph{et~al.}, ``Distributed optimization and statistical learning via the alternating direction method of multipliers,'' \emph{Foundations and Trends{\textregistered} in Machine Learning}, vol.~3, no.~1, pp. 1--122, 2011.

\bibitem{FastLio2}
W.~Xu, Y.~Cai, D.~He, J.~Lin, and F.~Zhang, ``Fast-lio2: Fast direct lidar-inertial odometry,'' \emph{IEEE Transactions on Robotics}, vol.~38, no.~4, pp. 2053--2073, 2022.

\bibitem{super}
Y.~Ren, F.~Zhu, G.~Lu, Y.~Cai, L.~Yin, F.~Kong, J.~Lin, N.~Chen, and F.~Zhang, ``Safety-assured high-speed navigation for mavs,'' \emph{Science Robotics}, vol.~10, no.~98, p. eado6187, 2025.

\end{thebibliography}

\end{document}